\documentclass[a4paper]{article}

\usepackage{ISCSLP2021}
\usepackage{multicol}
\title{Using heterogeneity in semi-supervised transcription hypotheses to improve code-switched speech recognition}
\name{Andrew Slottje, Shannon Wotherspoon, William Hartmann, Matthew Snover, Owen Kimball}
\address{Raytheon BBN Technologies, Cambridge, MA, USA}
\email{\{andrew.j.slottje, shannon.l.wotherspoon, william.hartmann, matt.snover, owen.kimball\} @raytheon.com}

\begin{document}

\maketitle
\begin{abstract}
 
 Modeling code-switched speech is an important problem in automatic speech recognition (ASR). Labeled code-switched data are rare, so monolingual data are often used to model code-switched speech. These monolingual data may be more closely matched to one of the languages in the code-switch pair. We show that such asymmetry can bias prediction toward the better-matched language and degrade overall model performance.
 
 
 To address this issue, we propose a semi-supervised approach for code-switched ASR. We consider the case of English-Mandarin code-switching, and the problem of using monolingual data to build bilingual ``transcription models'' for annotation of unlabeled code-switched data. 
 
 We first build multiple transcription models so that their individual predictions are variously biased toward either English or Mandarin. We then combine these biased transcriptions using confidence-based selection. This strategy generates a superior transcript for semi-supervised training, and obtains a 19\% relative improvement compared to a semi-supervised system that relies on a transcription model built with only the best-matched monolingual data.
 
 

\end{abstract}

\section{Introduction}

Multilingual speakers sometimes converse by code-switching, ``the spontaneous alternation of two languages in one stretch of discourse'' \cite{bakker}. In many multilingual societies, code-switching is a common form of communication \cite{scotton, hall}.


Domain match is a key issue in the modeling of code-switched speech. Because of the relative scarcity of transcribed code-switched speech data, modeling code-switching often requires the use of monolingual data. These monolingual datasets may be mismatched in a variety of ways to the code-switched target domain \cite{sitaram}. Some types of mismatch are intrinsic to the distinction between monolingual and code-switched data -- for example, n-gram language models built from combinations of monolingual corpora cannot encompass points of transition between languages, except under back-off. Other types of domain mismatch can include common ASR issues like training/test divergence in the register or dialect of the data. 



In response to such issues, some recent research has focused on the use of semi-supervised training, where unlabeled code-switched speech data are decoded and the hypotheses are used as pseudo-transcripts to build a supervised model \cite{guo,Biswas2019,yilmaz}. This line of inquiry invites the question of how to create an optimal ``transcription model'' for generating the annotations.

Y\i lmaz et al. \cite{yilmaz} consider the problem of Frisian-Dutch code-switching and propose an approach where language identification is used to select an optimal monolingual model for transcription. Biswas et al. \cite{Biswas2019} build bilingual models for four code-switch pairs in South African languages, then employ confidence scores to select an optimal transcription model. 

Like these approaches, our approach uses system combination to create an optimal transcription model from several systems. In particular, we consider the case of English-Mandarin code-switched speech, using monolingual English and Mandarin datasets as training data. 

Our datasets demonstrate various degrees of mismatch to our target domain, most notably accent mismatch. We show that when the training data for one language in a code-switch pair is more similar to the target domain than the training data for the other language, the resulting model disproportionately predicts in the higher-similarity language, making substitutions of English for Mandarin, or vice versa. We refer to this as the ``language confusability problem'' of code-switched ASR.


We leverage this finding to build multiple bilingual models, with some skewed toward Mandarin by the use of better-matched Mandarin training data, and others skewed toward English. We then combine these models using confidence-based selection, which helps neutralize the bias of the constituent systems. Combining systems in this way helps attack the language confusability problem, which appears otherwise intrinsic to code-switched domain adaptation.

Our strategy outperforms a baseline system that uses only well-matched data, as well as a baseline system that combines only monolingual models rather than bilingual ones. We also perform ablative work to show that these benefits accrue from system combination in particular, rather than from the inclusion of additional data or from dataset heterogeneity in our best model compared to our baseline.

While there is previous work where semi-supervised training is used for adaptation to a code-switched target domain \cite{guo, Biswas2019, yilmaz} as well as previous work leveraging dialectal heterogeneity in semi-supervised training \cite{dcq}, to our knowledge there has not been work on the use of the latter for code-switched ASR, nor demonstration of its special benefits in multilingual settings.

\section{Data}

The ``South-East Asia Mandarin-English'' (SEAME) corpus \cite{seame} includes data from speakers in Singapore and Malaysia engaged in Mandarin-English code-switching. It includes conversational speech as well as interview responses. The corpus includes transcriptions and a training-test split that provides for 103 hours of training data and 12 hours of test data.

Whereas previous semi-supervised research with this corpus has investigated the use of semi-supervised learning to supplement the transcribed labels \cite{guo}, we mimic the under-resourced nature of most code-switching problems by treating the SEAME corpus as a source of unlabeled data. We model these data with out-of-domain training sets, including Chinese-accented Mandarin (MAN), U.S.-accented English (UsEN), and Singaporean-accented English (SgEN). We sample where necessary to build three corpora of around 58 hours each from these datasets. Corpora are enumerated in Table \ref{t1}.


\begin{table}[th]
  \caption{Primary corpora.}
  \label{t1}
  \centering
  \begin{tabular}{ l@{}c  c }
    \toprule
    \multicolumn{1}{c}{\textbf{Description}} & \multicolumn{1}{c}{\textbf{Hrs Data}} & \multicolumn{1}{c}{\textbf{Label}} \\
    \midrule
    \multicolumn{3}{l}{\emph{Primary datasets.}}\\
    SEAME & 103 & SEAME\\
    Chinese Mandarin & 57 & MAN\\
    Singaporean English & 58 & SgEN\\
    U.S. English & 58 & UsEN\\
    \midrule
    \multicolumn{3}{l}{\emph{Secondary datasets}}\\
    SEAME English segments only & 23 & SmeE\\
    SEAME Mandarin segments only & 26 & SmeM\\
    SmeE+SmeM full utterances & 54 & SmeCS\\
    \bottomrule
  \end{tabular}
  
\end{table}

\section{Methods}

All ASR models presented in this paper are trained using the Kaldi toolkit \cite{kaldi} for acoustic modeling \cite{lfmmi,smbr} as in \cite{sage}, using the multilingual initialization procedure for neural network training elaborated in \cite{Ma2017}. We use a TDNN-F structure for the neural network architecture \cite{tdnnf}. We train this network for one epoch using the lattice-free MMI objective function \cite{lfmmi}, followed by two epochs using sMBR \cite{kingsbury}.

We use a trigram language model. In our bilingual language models, we assign each corpus an ``interpolation weight''; if not otherwise indicated, the weights used are uniform (i.e., $1/N$ for each corpus, where $N$ is the number of corpora used).

For semi-supervised acoustic model training, we use hypothesized pseudo-transcripts for the SEAME training set as labels for one epoch of lattice-free MMI training alongside supervised data, followed by two fully supervised sMBR epochs.


We report results for error rate measured on English, Mandarin, and code-switched utterances, as well as an overall error rate. Mandarin predictions are scored at the character level (``CER''), while English predictions are scored at the word level (``WER''). The reported ``mixed error rate'' (``MER'') aggregates these metrics.


\section{Domain mismatch with code-switching}

The Mandarin and English varieties spoken on the Malay Peninsula can differ materially from higher-resource varieties like Standard Chinese and General American English. For example, tones in Singapore Mandarin can differ from those of Standard Chinese \cite{sgtones}, and Singapore English sometimes lacks certain vowel length distinctions observed in U.S. English \cite{wee}. For such reasons, domain-shift issues may arise when using American or Chinese speech data to model Malaysian or Singaporean speech.


In all our experiments, we orient ourselves with respect to the SEAME target domain of English-Mandarin code-switching in Malaysia and Singapore. Training on U.S. English data (for example) therefore introduces what we will refer to by the general term of ``domain mismatch,'' encompassing such disparities as these potential accent differences, as well as issues like potential differences in channel characteristics. When we train on the SEAME dataset or its subsets, we refer to the training data used as ``domain-matched.'' We also use the terms ``in-domain'' and ``out-of-domain'' in a similar manner.
\begin{table}[th]
  \caption{Code-switched data provide only a small improvement over domain-matched monolingual data.}
  \label{t2}
  \centering
  \begin{tabular}{c c c c l l}
    \toprule
    \multicolumn{1}{c}{\textbf{CER}} & \multicolumn{1}{c}{\textbf{WER}} & \multicolumn{1}{c}{\textbf{MER}} & \multicolumn{1}{c}{\textbf{MER}} & \multicolumn{1}{c}{\textbf{AM Data}} & \multicolumn{1}{c}{\textbf{LM Data}} \\
        \multicolumn{1}{c}{\textbf{(M)}} & \multicolumn{1}{c}{\textbf{(E)}} & \multicolumn{1}{c}{\textbf{(CS)}} & \multicolumn{1}{c}{\textbf{(All)}} & \multicolumn{1}{c}{} & \multicolumn{1}{c}{} \\
    \midrule
    27.6 & 31.2 & 25.9 & 27.0 & SmeCS & SmeCS\\
    27.4 & 31.3 & 26.6 & 27.5 & SmeCS & SmeM+\\
    &&&&& SmeE\\
    27.8 & 31.7 & 27.2 & 28.0 & SmeM+ & SmeCS\\
    &&&& SmeE &\\
    27.9 & 31.7 & 28.0 & 28.7 & SmeM+ & SmeM+\\
    &&&& SmeE & SmeE\\
    \bottomrule
  \end{tabular}
  \end{table}

We perform experiments that show that domain-matched monolingual data are broadly sufficient for modeling a code-switched target domain, but that domain-mismatched monolingual data produce large degradations in performance. We argue that, taken together, these results indicate a central role for the problem of domain mismatch, especially in acoustic modeling.

We first create an artificial monolingual dataset from the SEAME data and show that these monolingual data are sufficient to model the code-switching target domain. We use GMM-based time alignments of the SEAME training data to split utterances at language boundaries, into new Mandarin and English utterances which must meet a length criterion. If any new utterance is too short, none are retained. From these new utterances we constitute 23-hour monolingual English (SmeE) and Mandarin (SmeM) corpora. We also build a 54-hour SmeCS corpus from the unsplit antecedents of the utterances that are retained in SmeE and SmeM. We present results from models trained on these data in Table \ref{t2}. Using matched code-switching data instead of matched monolingual data contributes only 2.1 percentage points MER improvement to performance on code-switched utterances, and only 1.7 percentage points MER on the whole test set.



\begin{table}[th]
  \caption{Acoustic model domain mismatch causes large performance degradation.}
  \label{t3}
  \centering
  \begin{tabular}{c c c c l l}
    \toprule
    \multicolumn{1}{c}{\textbf{CER}} & \multicolumn{1}{c}{\textbf{WER}} & \multicolumn{1}{c}{\textbf{MER}} & \multicolumn{1}{c}{\textbf{MER}} & \multicolumn{1}{c}{\textbf{AM Data}} & \multicolumn{1}{c}{\textbf{LM Data}} \\
        \multicolumn{1}{c}{\textbf{(M)}} & \multicolumn{1}{c}{\textbf{(E)}} & \multicolumn{1}{c}{\textbf{(CS)}} & \multicolumn{1}{c}{\textbf{(All)}} & \multicolumn{1}{c}{} & \multicolumn{1}{c}{} \\
    \midrule
    26.6 & 31.0 & 23.6 & 25.2 & SEAME & SEAME\\
    58.1 & 38.8 & 50.3 & 49.2 & MAN+ & SEAME\\
    &&&& SgEN &\\
    61.5 & 44.8 & 56.5 & 55.0 & MAN+ & SEAME\\
    &&&& UsEN &\\
    \bottomrule
  \end{tabular}
  \end{table}
  
\begin{table}[th]
  \caption{Language model domain mismatch causes relatively minor increase in error rates.}
  \label{t4}
  \centering
  \begin{tabular}{c c c c l l}
    \toprule
    \multicolumn{1}{c}{\textbf{CER}} & \multicolumn{1}{c}{\textbf{WER}} & \multicolumn{1}{c}{\textbf{MER}} & \multicolumn{1}{c}{\textbf{MER}} & \multicolumn{1}{c}{\textbf{AM Data}} & \multicolumn{1}{c}{\textbf{LM Data}} \\
        \multicolumn{1}{c}{\textbf{(M)}} & \multicolumn{1}{c}{\textbf{(E)}} & \multicolumn{1}{c}{\textbf{(CS)}} & \multicolumn{1}{c}{\textbf{(All)}} & \multicolumn{1}{c}{} & \multicolumn{1}{c}{} \\
    \midrule
    26.6 & 31.0 & 23.6 & 25.2 & SEAME & SEAME\\
    27.9 & 38.2 & 28.0 & 29.8 & SEAME & MAN+\\
    &&&&& SgEN\\
    27.9 & 38.2 & 27.7 & 29.5 & SEAME & MAN+\\
    &&&&& UsEN\\
    \bottomrule
  \end{tabular}
  \end{table}
  
By comparison, using training data that are poorly matched to the SEAME target domain sharply degrades performance. The impact of these differences is most pronounced in acoustic modeling, consistent with the role we attribute to characteristics such as accent differences, while mismatch in language data appears relatively unimportant. Table \ref{t3} shows that using out-of-domain data for acoustic modeling can double the error rate of our model. Table \ref{t4} shows a relatively minor degradation from using the same out-of-domain monolingual data for language modeling.


\begin{table}[th]
  \caption{Differences in training-test domain match between the languages of a code-switch pair degrade performance in the more poorly matched language.}
  \label{t5}
  \centering
  \begin{tabular}{c c c c l}
    \toprule
    \multicolumn{1}{c}{\textbf{CER}} & \multicolumn{1}{c}{\textbf{WER}} & \multicolumn{1}{c}{\textbf{MER}} & \multicolumn{1}{c}{\textbf{MER}} & \multicolumn{1}{c}{\textbf{Training Data}} \\
        \multicolumn{1}{c}{\textbf{(M)}} & \multicolumn{1}{c}{\textbf{(E)}} & \multicolumn{1}{c}{\textbf{(CS)}} & \multicolumn{1}{c}{\textbf{(All)}} & \multicolumn{1}{c}{}  \\
    \midrule
    26.6 & 31.0 & 23.6 & 25.2 & SEAME \\
    \midrule
    \multicolumn{5}{l}{\emph{Results with mismatched English:}}\\
    61.8 & 44.9 & 56.3 & 54.9 & MAN+SgEN\\
    48.9 & 73.2 & 54.6 & 57.2 & MAN+UsEN\\
    28.5 & 90.4 & 47.4 & 52.7 & SmeM+SgEN\\
    28.7 & 99.5 & 50.4 & 56.5 & SmeM+UsEN\\
    \midrule
    \multicolumn{5}{l}{\emph{Results with mismatched Mandarin:}}\\
    61.8 & 44.9 & 56.3 & 54.9 & MAN+SgEN\\
    48.9 & 73.2 & 54.6 & 57.2 & MAN+UsEN\\
    92.6 & 31.2 & 72.1 & 67.3 & MAN+SmeE\\
    \bottomrule
  \end{tabular}
  \end{table}

\begin{table}[th]
  \caption{Close training-test domain match for only one language in a code-switch pair causes cross-language substitution.}
  \label{t6}
  \centering
  \begin{tabular}{c c l}
    \toprule
    \multicolumn{1}{c}{\textbf{\% Eng hyp}} & \multicolumn{1}{c}{\textbf{\% Man hyp}} & \multicolumn{1}{c}{\textbf{Training Data}}\\
        \multicolumn{1}{c}{\textbf{as Man}} & \multicolumn{1}{c}{\textbf{as Eng}} &  \multicolumn{1}{c}{} \\
    \midrule
    3.2 & 2.4 & SEAME \\
    \midrule
    2.9 & 32.8 & MAN+SgEN \\
    26.8 & 11.9 & MAN+UsEN \\
    60.7 & 0.0 & SmeM+SgEN \\
    67.7 & 0.0 & SmeM+UsEN \\
    0.0 & 47.8 & MAN+SmeE \\
    \bottomrule
  \end{tabular}
  \end{table}

Tables \ref{t5} and \ref{t6} explain the central role of domain mismatch by presenting evidence of its relationship to language confusability. We find that narrowing the domain shift between only one of the training datasets and the target domain – i.e., picking a more similar English or Mandarin training set, while holding the other constant – naturally improves test performance on the language with better-matched data. But surprisingly, performance on the other language also sharply degrades: if a model has perfectly matched Mandarin training data but poorly matched English training data, it will rarely predict English. 

\section{A code-switching optimized approach for semi-supervised training}

\subsection{Background}
To attack the language confusability problem, we build several systems with varying degrees of English and Mandarin match to the SEAME target domain, and we use confidence-based model selection to select a prediction from among these. This allows us to neutralize the propensity of some bilingual models to overpredict in English or Mandarin, while leveraging bilingual training data to improve predictions and confidence estimates. 

This approach yields superior hypotheses with less cross-language substitution, which we use as pseudo-transcripts for semi-supervised training. We compare a system trained in this manner with systems where the pseudo-transcripts are generated by only one of the bilingual models, without the bias-neutralizing aid of the system combination. Compared to such systems, our combined system demonstrates better implicit language recognition capabilities and produces a generally superior ASR system from the use of more accurate training data. 
	
For the system combination, we make use of the ROVER ``maximum confidence scores'' method \cite{rover}.

\subsection{Baseline model}

Combining the MAN+SgEN and the MAN+UsEN models is the key to our approach. Using ROVER to join these models greatly improves prediction, yielding relative improvements of over 15\% on Mandarin (compared to MAN+SgEN) and of over 30\% on English (compared to MAN+UsEN). Tellingly, the combined model's performance on code-switched utterances also demonstrates great improvement over either of the constituent models. Results are in Tables \ref{t9} and \ref{t10}.

The combined system additionally outperforms combinations of monolingual models (compare scores for ROVER\{A,B\} to ROVER\{R1,R2\} and ROVER 3 in Table \ref{t9}). We hypothesize this may be due to additional information in the confidence estimates of the bilingual models.


\begin{table}[th]
  \caption{Index of ROVER systems.}
  \label{t8}
  \centering
  \begin{tabular}{c c l}
    \toprule
    \multicolumn{1}{c}{\textbf{Index}} & \multicolumn{1}{c}{\textbf{Systems}} \\
    \midrule
    ROVER 1 & MAN+SgEN, MAN+UsEN\\
    ROVER 2 & \{MAN+SgEN, MAN+UsEN\}$\times$\\
    & {LM Interpolation Weight Variation}\\
    ROVER 3 & MAN, SgEN, UsEN\\
    ROVER 3A & MAN, SgEN\\
    ROVER 3B & MAN, UsEN\\
    ROVER 4 & ROVER 1 + ROVER 3\\
    ROVER 5 & ROVER 2 + ROVER 3\\
    \bottomrule
  \end{tabular}
  \end{table}

\begin{table}[th]
  \caption{Combining bilingual models outperforms combinations of monolingual models. Systems are indexed in Table \ref{t8}.}
  \label{t9}
  \centering
  \begin{tabular}{c c c c c l}
    \toprule
    \multicolumn{1}{c}{\textbf{ID}} & \multicolumn{1}{c}{\textbf{CER}} & \multicolumn{1}{c}{\textbf{WER}} & \multicolumn{1}{c}{\textbf{MER}} & \multicolumn{1}{c}{\textbf{MER}} & \multicolumn{1}{c}{\textbf{Training Data}} \\
        \multicolumn{1}{c}{} & \multicolumn{1}{c}{\textbf{(M)}} & \multicolumn{1}{c}{\textbf{(E)}} & \multicolumn{1}{c}{\textbf{(CS)}} & \multicolumn{1}{c}{\textbf{(All)}} & \multicolumn{1}{c}{}  \\
    \midrule
    A & 61.8 & 44.9 & 56.3 & 54.9 & MAN+SgEN\\
    B & 48.9 & 73.2 & 54.6 & 57.2 & MAN+UsEN\\
    & 51.0 & 47.5 & 48.0 & 48.2 & ROVER\{A,B\}\\
    R1 & 55.9 & 50.6 & 52.6 & 52.6 & ROVER 3A\\
    R2 & 54.4 & 73.9 & 59.2 & 61.2 & ROVER 3B\\
    & 63.2 & 53.8 & 59.5 & 58.9 & ROVER\{R1,R2\}\\
    & 50.7 & 60.2 & 51.7 & 53.1 & ROVER 3\\
    \bottomrule
  \end{tabular}
  \end{table}
  
  \begin{table}[th]
  \caption{The proposed system combination method diminishes bias in cross-language substitution.}
  \label{t10}
  \centering
  \begin{tabular}{c c c l}
    \toprule
    \multicolumn{1}{c}{\textbf{ID}} & \multicolumn{1}{c}{\textbf{\% Eng hyp}} & \multicolumn{1}{c}{\textbf{\% Man hyp}} & \multicolumn{1}{c}{\textbf{Training Data}}\\
        \multicolumn{1}{c}{} & \multicolumn{1}{c}{\textbf{as Man}} & \multicolumn{1}{c}{\textbf{as Eng}} &  \multicolumn{1}{c}{} \\
    \midrule
    & 3.2 & 2.4 & SEAME \\
    \midrule
    A & 2.9 & 32.8 & MAN+SgEN\\
    B & 26.8 & 11.9 & Man+UsEN\\
    & 6.6 & 19.3 & ROVER\{A,B\}\\
    \bottomrule
  \end{tabular}
  \end{table}
  
Table \ref{t11} demonstrates that these benefits are due principally to system combination, rather than to the availability of heterogeneous data in training. Compared to the MAN+SgEN and MAN+UsEN baselines, using both Singapore English and U.S. English in training only modestly improves English performance, while it markedly degrades Mandarin performance beyond that of the MAN+SgEN or MAN+UsEN models.

\begin{table}[th]
  \caption{System combination outperforms training on heterogeneous data all at once. SgEN28 and UsEN28 are 28-hour samples from SgEN and UsEN to hold dataset size constant.}
  \label{t11}
  \centering
  \begin{tabular}{c c c c l}
    \toprule
    \multicolumn{1}{c}{\textbf{CER}} & \multicolumn{1}{c}{\textbf{WER}} & \multicolumn{1}{c}{\textbf{MER}} & \multicolumn{1}{c}{\textbf{MER}} & \multicolumn{1}{c}{\textbf{Training Data}} \\
        \multicolumn{1}{c}{\textbf{(M)}} & \multicolumn{1}{c}{\textbf{(E)}} & \multicolumn{1}{c}{\textbf{(CS)}} & \multicolumn{1}{c}{\textbf{(All)}} & \multicolumn{1}{c}{}  \\
    \midrule
    26.6 & 31.0 & 23.6 & 25.2 & SEAME \\
    \midrule
    61.8 & 44.9 & 56.3 & 54.9 & MAN+SgEN\\
    48.9 & 73.2 & 54.6 & 57.2 & MAN+UsEN\\
    51.0 & 47.5 & 48.0 & 48.2 & ROVER 1\\
    \midrule
    76.5 & 43.6 & 64.9 & 62.5 & MAN+SgEN28+\\
    &&&& UsEN28\\
    \bottomrule
  \end{tabular}
  \end{table}

\subsection{Transcription model improvement}
We develop further improvements to the transcription model beyond those of Section 5.2. We do so by expanding the system combination to include interpolation weight variation in the combined systems, as well as fully monolingual models. 

``Interpolation weight,'' as discussed in Section 3, refers to the weight we give to a particular corpus in our language model. In our setting, varying the interpolation weight has the effect of biasing the language model toward a particular language. Combining our original baseline models with models where English is up-weighted and English performance is better, as well as with models where Mandarin is up-weighted and Mandarin performance is better, yields improved performance in both English and Mandarin. The strategy at work here -- combining bilingual models that are biased toward English or Mandarin to improve recognition in both languages -- mirrors our central scheme discussed in Section 5.2. 

Results are in Table \ref{t12}. Each of these improvements is helpful individually, and together they boost transcription model performance by 2\% MER.


\begin{table}[th]
  \caption{Transcription model error rates on SEAME test set improve under system combination.}
  \label{t12}
  \centering
  \begin{tabular}{c c c c l}
    \toprule
    \multicolumn{1}{c}{\textbf{CER}} & \multicolumn{1}{c}{\textbf{WER}} & \multicolumn{1}{c}{\textbf{MER}} & \multicolumn{1}{c}{\textbf{MER}} & \multicolumn{1}{c}{\textbf{Training Data}} \\
        \multicolumn{1}{c}{\textbf{(M)}} & \multicolumn{1}{c}{\textbf{(E)}} & \multicolumn{1}{c}{\textbf{(CS)}} & \multicolumn{1}{c}{\textbf{(All)}} & \multicolumn{1}{c}{}  \\
    \midrule
    61.8 & 44.9 & 56.3 & 54.9 & MAN+SgEN\\
    48.9 & 73.2 & 54.6 & 57.2 & MAN+UsEN\\
    51.0 & 47.5 & 48.0 & 48.2 & ROVER 1\\
    49.7 & 47.5 & 47.0 & 47.4 & ROVER 2\\
    50.7 & 60.2 & 51.7 & 53.1 & ROVER 3\\
    48.1 & 49.3 & 46.5 & 47.2 & ROVER 4\\
    47.8 & 47.7 & 45.6 & 46.2 & ROVER 5\\
    \bottomrule
  \end{tabular}
  \end{table}

\begin{table}[th]
  \caption{Transcription model improvements from system combination improve performance of semi-supervised models.}
  \label{t13}
  \centering
  \begin{tabular}{c c c c l l}
    \toprule
    \multicolumn{1}{c}{\textbf{CER}} & \multicolumn{1}{c}{\textbf{WER}} & \multicolumn{1}{c}{\textbf{MER}} & \multicolumn{1}{c}{\textbf{MER}} & \multicolumn{1}{c}{\textbf{Training Data}} & \multicolumn{1}{c}{\textbf{SST}} \\
        \multicolumn{1}{c}{\textbf{(M)}} & \multicolumn{1}{c}{\textbf{(E)}} & \multicolumn{1}{c}{\textbf{(CS)}} & \multicolumn{1}{c}{\textbf{(All)}} & \multicolumn{1}{c}{} & \multicolumn{1}{c}{\textbf{AM/LM}} \\
    \midrule
    62.0 & 42.2 & 54.4 & 53.1 & MAN+SgEN & AM \\
    54.5 & 42.6 & 48.5 & 48.1 & MAN+SgEN & AM+LM\\
    51.8 & 65.0 & 53.1 & 55.0 & MAN+UsEN & AM\\
    42.3 & 67.5 & 47.7 & 50.6 & MAN+UsEN & AM+LM\\
    55.5 & 40.6 & 47.2 & 47.0 & ROVER 1 & AM\\
    45.9 & 41.4 & 40.1 & 41.0 & ROVER 1 & AM+LM\\
    54.5 & 40.3 & 46.5 & 46.3 & ROVER 2 & AM\\
    44.1 & 41.1 & 39.0 & 39.9 & ROVER 2 & AM+LM\\
    52.1 & 40.7 & 45.1 & 45.1 & ROVER 4 & AM\\
    41.9 & 41.7 & 38.1 & 39.1 & ROVER 4 & AM+LM\\
    52.2 & 40.4 & 45.1 & 45.1 & ROVER 5 & AM\\
    41.7 & 41.2 & 38.0 & 39.0 & ROVER 5 & AM+LM\\
    \bottomrule
  \end{tabular}
  \end{table}

\subsection{Semi-supervised results}

Our results in Table \ref{t13} demonstrate that the transcription model improvements of Sections 5.2 and 5.3 carry over into the semi-supervised domain. 


Using system combination to generate the pseudo-transcripts produces an improvement in both acoustic and language modeling, in line with the improvements shown in Tables \ref{t9} and \ref{t12}. The simplest system combination we test (MAN+SgEN combined with MAN+UsEN) yields a 7.1 percent absolute MER improvement on our test set compared to a model trained with MAN+SgEN-only transcription. This represents a nearly 15\% relative improvement over the MAN+SgEN baseline. The improvement is even larger when a model trained with the MAN+UsEN-only transcription is used as the point of comparison.

The best transcription system (ROVER 5) supplements the baseline system combination with input from multiple language model interpolation weight settings and monolingual models as in Section 5.3. This system yields a 9.1 percent MER improvement over the baseline system, a 19\% relative improvement.

\section{Conclusions}

Out-of-domain data are frequently used of necessity for ASR modeling of code-switched speech. Due to the fact that code-switching often occurs into a standard register or lingua franca, the primary language of a code-switched conversation may be a lower-resource dialect or accent, with concomitant limitations in availability of well-matched monolingual data \cite{auer}. We expect that the domain match issues we have identified are common to many of the settings where code-switching ASR has been attempted. 

In such a setting, heterogeneous data may be available for the secondary language in the code-switch pair, making it possible to consider an approach similar to the one we have taken here. We believe our insights and approach may be of interest in such situations.





\bibliographystyle{IEEEtran}

\bibliography{mybib}


\end{document}